\documentclass[10pt, conference]{IEEEtran}

\usepackage{amsmath,amssymb,amsfonts}
\usepackage{graphicx}
\usepackage{pifont}
\usepackage{algorithmic}

\usepackage{amssymb}
\usepackage{amsthm}
\usepackage{xcolor}
\usepackage[figuresright]{rotating}

\usepackage{graphicx}
\graphicspath{{/}{fig/}}

\usepackage{array}
\usepackage{textcomp}
\usepackage{gensymb}
\usepackage{xcolor}

\usepackage{mathtools}
\usepackage[vlined, ruled, shortend]{algorithm2e}

\usepackage[style=base]{caption}
\usepackage{subcaption}

\usepackage{float}
\usepackage{hyperref}
\usepackage{enumerate}

\usepackage{pgfplots}
\pgfplotsset{compat=1.7}
\usepgfplotslibrary{groupplots}
\usetikzlibrary{matrix}

\usepackage[ancient]{flushend}

\newcommand{\RN}[1]{%
  \textup{\uppercase\expandafter{\romannumeral#1}}%
}

\usepackage{xcolor}
\colorlet{orange_maxim}{green!10!orange!90!}

\usepackage[capitalise]{cleveref}

\newlength\figureheight
\newlength\figurewidth
\usetikzlibrary{shapes.misc}

\tikzset{cross/.style={cross out, draw=black, minimum size=2*(#1-\pgflinewidth), inner sep=0pt, outer sep=0pt},
cross/.default={1pt}}

\renewcommand{\baselinestretch}{0.985}

\IEEEaftertitletext{\vspace{-1.5\baselineskip}}

\title{
    Adaptive Lidar Scan Frame Integration:\\
    Tracking Known MAVs in 3D Point Clouds
}

\author{
    \IEEEauthorblockN{
        \vspace{1em}
        Li Qingqing\IEEEauthorrefmark{1},
        Yu Xianjia\IEEEauthorrefmark{1},
        Jorge Pe\~na Queralta\IEEEauthorrefmark{1},
        Tomi Westerlund\IEEEauthorrefmark{1} \\[+0.5em]
    }
    \IEEEauthorblockA{
        \normalsize
        \IEEEauthorrefmark{1}\href{https://tiers.utu.fi}{Turku Intelligent Embedded and Robotic Systems (TIERS) Lab, University of Turku, Finland}.\\
        Emails: \textsuperscript{1}\{qingqli, xianjia.yu, jopequ, tovewe\}@utu.fi\\[+6pt]
    }
}

\begin{document}

\maketitle
\thispagestyle{empty}
\pagestyle{empty}


\begin{abstract}
    
    %
    Micro-aerial vehicles (MAVs) are becoming ubiquitous across multiple industries and application domains. Lightweight MAVs with only an onboard flight controller and a minimal sensor suite (e.g., IMU, vision, and vertical ranging sensors) have potential as mobile and easily deployable sensing platforms. When deployed from a ground robot, a key parameter is a relative localization between the ground robot and the MAV.
    This paper proposes a novel method for tracking MAVs in lidar point clouds. In lidar point clouds, we consider the speed and distance of the MAV to actively adapt the lidar's frame integration time and, in essence, the density and size of the point cloud to be processed. We show that this method enables more persistent and robust tracking when the speed of the MAV or its distance to the tracking sensor changes. 
    In addition, we propose a multi-modal tracking method that relies on high-frequency scans for accurate state estimation, lower-frequency scans for robust and persistent tracking, and sub-Hz processing for trajectory and object identification. These three integration and processing modalities allow for an overall accurate and robust MAV tracking while ensuring the object being tracked meets shape and size constraints.
 

\end{abstract}

\IEEEpeerreviewmaketitle


\section{Introduction}\label{sec:introduction}


Micro-aerial vehicles (MAVs) have seen an increasing adoption across a variety of application domains in recent years~\cite{blosch2010vision}. Multiple works have been devoted to the navigation of MAVs in GNSS-denied environments~\cite{nieuwenhuisen2016autonomous}, and state estimation in both single~\cite{song2017multi} and multi-MAV systems~\cite{xu2020decentralized}. In this paper, we are particularly interested in tracking and state estimation from an external system, for those applications where MAVs are deployed together with or from unmanned ground vehicles (UGVs)~\cite{gawel2018aerial, queralta2020collaborative}.

From the perspective of deployment within multi-robot systems, being able to track MAVs from UGVs enables miniaturization and higher degrees of flexibility lowering the need for high-accuracy onboard localization. A recent and significant example of multi-robot system deployment in GNSS-denied environments is the DARPA Subterranean challenge~\cite{rouvcek2019darpa, petrlik2020robust}. Reports from participating teams indicate that localization and collaborative sensing were among the key challenges, with MAVs being deployed from UGVs dynamically during the challenge. Since MAVs often rely on visual-inertial odometry (VIO) for self and relative estate estimation~\cite{nguyen2020vision}, relying on external lidar-based tracking can also extend the operability to low-visibility or other domains where VIO has inherent limitations~\cite{queralta2020vio, qingqing2019monocular}.

\begin{figure}
    \centering
    \begin{subfigure}{0.48\textwidth}
        \centering
        \includegraphics[width=\textwidth]{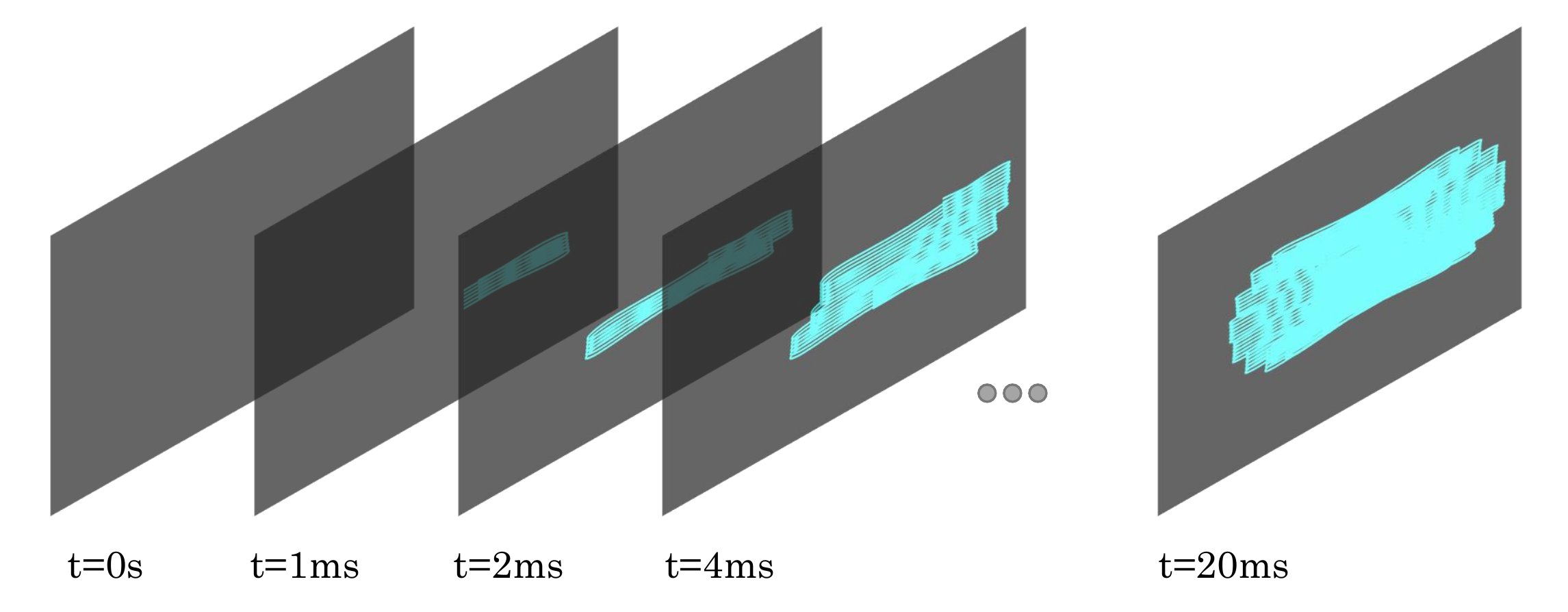}
        \caption{Illustration of the field of view (FoV) coverage with different point cloud integration times in a non-repetitive lidar scanning device.\\}
        \label{fig:concept_integration}
    \end{subfigure}
    \begin{subfigure}{0.48\textwidth}
        \centering
        \includegraphics[width=\textwidth]{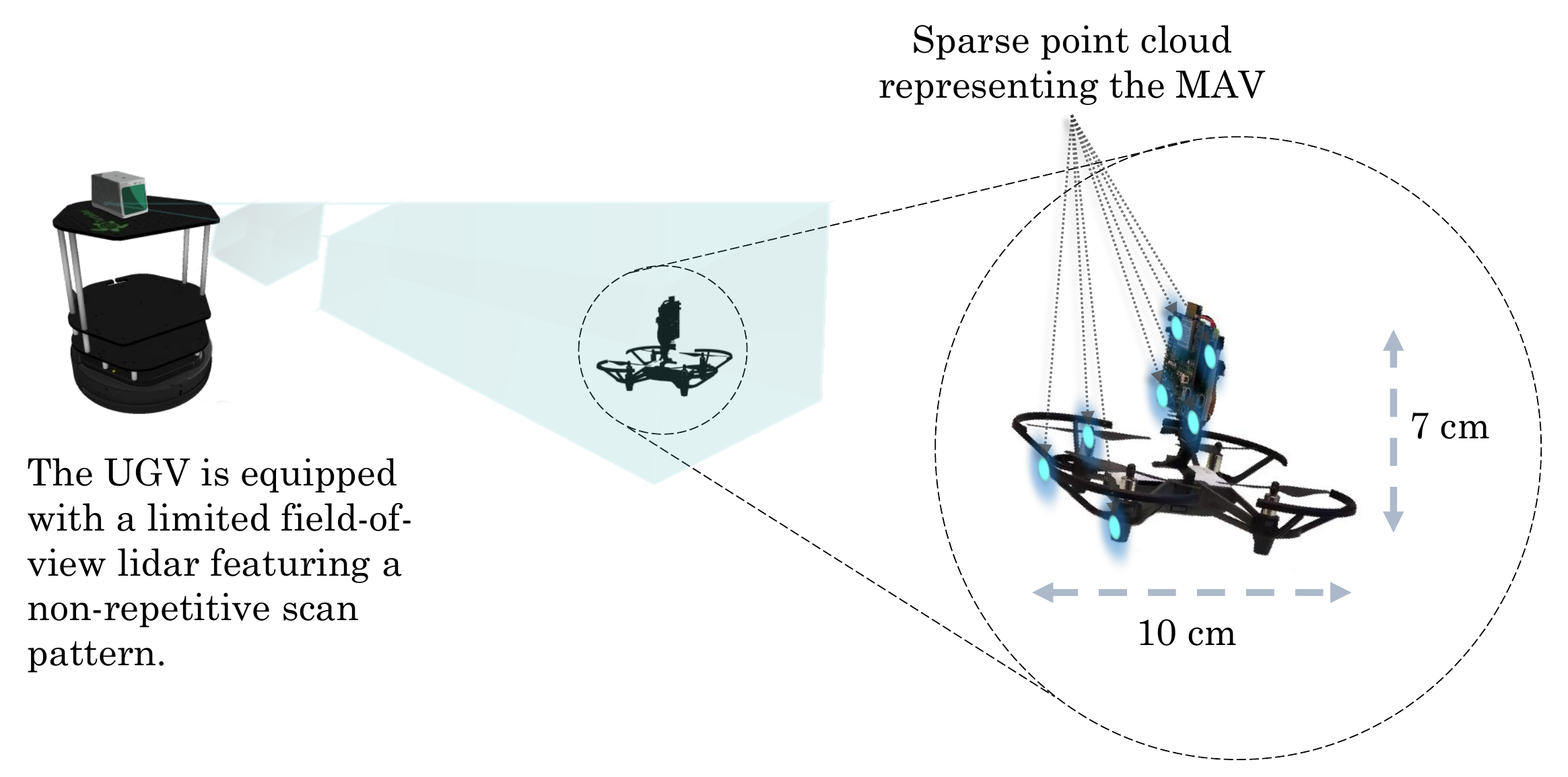}
        \caption{Illustration of a ground robot tracking a micro-aerial vehicle (MAV) using a limited-FoV solid-state lidar.}
        \label{fig:concept_tracking}
    \end{subfigure}
    \caption{Conceptual illustration of the field-of-view coverage with different integration times on a Livox Horizon lidar (top) and its application to tracking MAVs (bottom).}
    \vspace{1em}
    \label{fig:concept}
\end{figure}
 
Tracking and detecting MAVs has been a topic of interest for researchers in recent years. First, owing to the increasing need of identifying and detecting foreign objects or drones in areas with controlled airspace such as airports~\cite{guvenc2018detection, hengy2017multimodal}. Second, to optimize the utilization of MAVs as flexible mobile sensing platforms~\cite{queralta2020autosos}. This paper focuses on the latter use. Compared to the existing literature, which relies mainly on vision-based techniques~\cite{vrba2020marker}, we provide a lidar-based solution that can be utilized more independently of the environmental conditions. Until recently, most 3D lidars provided relatively sparse point clouds in terms of object recognition~\cite{razlaw2019detection}, with limited vertical resolution in inexpensive devices. However, solid-state lidars have recently emerged as state-of-the-art in terms of long-range scanners featuring high-density point clouds~\cite{li2020towards}. The main caveat is the limited field of view (FoV) in most of these devices~\cite {lin2020loam}, but solutions include utilizing multiple lidars or correspondingly adjusting the position and orientation of the robot base where the lidar is installed.
 
We are particularly interested in the problem of tracking a MAV that is deployed from a ground robot. We assume thus that the initial position of the MAV after take-off is known. We also assume that its shape and size are known a priori. We develop methods targeting solid-state lidars owing to the higher density of the resulting point cloud even with more limited FoV. Moreover, in these lidars, the concept of a frame or scan frequency changes considerably. Similarly, as in rotating 3D lidars, a frame in solid-state lidars can be naturally related to a single revolution. With non-repetitive scan patterns, lidars can output point clouds at adjustable frequencies with varying FoV coverage, as illustrated in Fig.~\ref{fig:concept_integration}. This opens the door to new lidar perception methods that exploit the possibilities of adaptively adjust the frame integration time to better sense the objects. To the best of our knowledge, this approach has not been previously studied. We apply the proposed adaptive lidar scan integration methods within the problem of a UGV tracking a MAV for external state estimation, as conceptualized in Fig.~\ref{fig:concept_tracking}. While our focus is on MAVs, the proposed methods can also be easily adapted to detect foreign objects or intruder MAVs more accurately. We first put our focus on single and known MAV detection, but present generic methods that can be extended to multi-MAV tracking as long as FoV limitations are accounted for.

The main contribution of this paper is twofold. We first introduce a novel adaptive lidar scan integration method enabling more accurate and reliable object recognition and tracking from 3D point clouds, specifically applied to MAV detection. In addition, we then define a multi-modal tracking system that relies on processing point clouds resulting in different integration times for higher accuracy and persistent tracking, while validating the trajectories using a priori known information about the MAV dimensions.

The remaining of this paper is organized as follows. In section II, we review the state-of-the-art in MAV detection, lidar-based object detection and tracking, and a handful of existing works on the vision-, radar- and lidar-based MAV detection and tracking. Section III then formulates the adaptive scanning method, and how it applies to a MAV detection and tracking problem. Section IV reports on our methodology, and Section V describes experimental results with different settings. Finally, we conclude this work and outline future research directions in Section VI.

\section{Related Works}\label{sec:related_works}

This section reviews the literature in the areas of detection and tracking of MAVs. Owing to the scarcity of works devoted to lidar-based MAV tracking, we have focused on: (i) the state-of-the-art in MAV detection, mostly vision-based; (ii) lidar-based detection and tracking of small objects; and (iii) detection of MAVs based on lidar or radar point cloud data.

\subsection{Vision-based MAV Detection}

Most of the work to data in tracking small objects and MAVs has been related to vision-based approaches~\cite{mueller2016benchmark, nguyen2017remote, vrba2020marker}. Vision-based approaches can be classified among those that rely on passive or active visual markers, and those that detect and track objects in general, e.g., with traditional computer vision or deep learning. In the former category, \cite{nguyen2017remote} provides an example of tracking based on passive artificial visual markers, which can be used to calculate the relative 3D position of the MAV from a camera. On a different direction aimed at MAV-to-MAV detection, Walter et al. presented UVDAR, an ultra-violet (UV) solution for relative localization in multi-MAV systems~\cite{walter2019uvdar}.

Regarding the latter category, the development of deep convolutional neural networks (CNNs) in recent years has facilitated the adoption within the domain of object detection and tracking. Arguably, a significant portion of the state-of-the-art in tracking is based on deep learning methods~\cite{li2018deep}. These methods often offer significantly higher degrees of accuracy and robustness. For instance, Vrba et al. have presented a marker-less system for relative localization of MAVs~\cite{vrba2020marker}, which can be applied to detecting foreign or intruder MAVs.

The potential of depth cameras for detecting MAVs has also been showcased in the literature. For instance, deep learning models processing depth maps have been applied to tracking a MAV and aiding it in navigating and avoiding obstacles~\cite{carrio2018drone}.

While depth cameras can provide accurate location and size measurements, and vision sensors, in general, are able of robust tracking and relative positioning, our focus in this paper is to work with lidars owing to their flexibility in terms of environmental conditions, and because of their significantly higher range and accuracy when compared to depth cameras.




\subsection{Lidar-based object tracking}

More in line with the research presented in this paper is point-cloud-based tracking. While this generally refers to lidar point clouds, some of the work in the literature is also devoted to point clouds generated by stereo or depth cameras, or radars. In general terms, traditional approaches to tracking in point cloud data rely mostly on distance-based clustering ~\cite{rangesh2019no}. 

Nonetheless, significant work has been carried out in the area of deep learning voxel-based methods for segmentation and detection of objects in 3D point clouds. For instance, VoxelNet~\cite{zhou2018voxelnet} implements a voxel features extractor (VFE) on point cloud to characterize object points. Other networks have been proposed that directly process point sets, such as PointNet~\cite{qi2017pointnet} and PointNet++~\cite{qi2017pointnet++}, to fully exploit the inherent information in the point cloud data for object tracking. Iterating over these, works such as~\cite{qi2020p2b} have proposed end-to-end and point-to-box networks for 3D object tracking.

When considering small objects, the specific literature is more scarce. In~\cite{razlaw2019detection}, Razlaw et al. focus on detecting people in sparse point clouds from multi-channel rotating 3D lidars. Compared with this approach, we focus on exploiting the adaptive frame integration capabilities of solid-state lidars to optimize the point cloud density and do not necessarily assume sparsity. 






\begin{figure*}
    \centering
    \includegraphics[width=\textwidth]{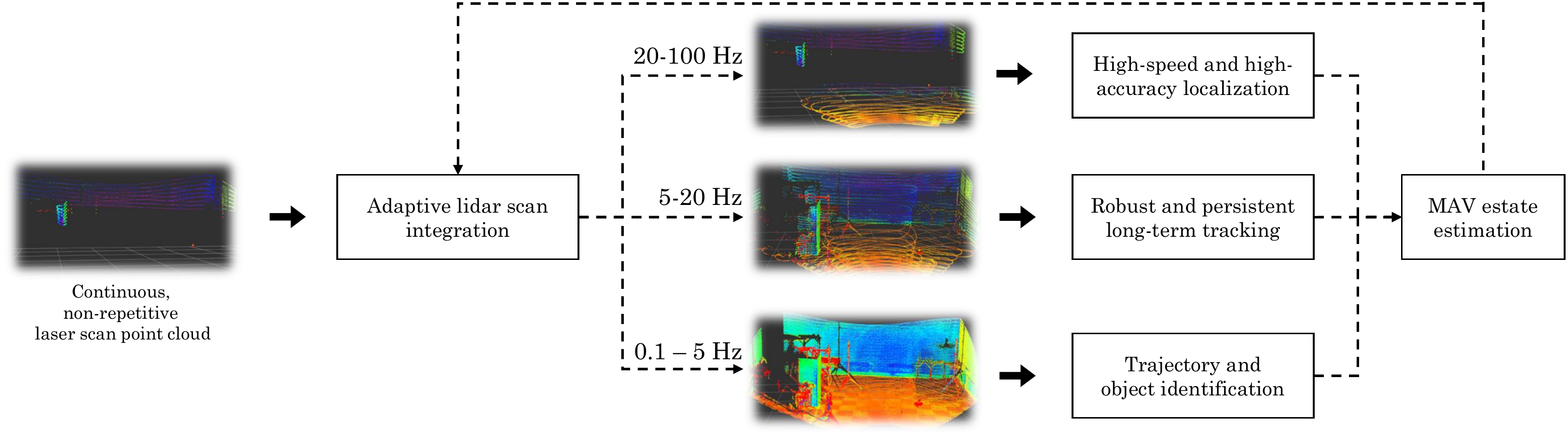}
    \caption{Overview of the proposed methods, where tracking is simultaneously performed at three different scan frequencies. Within each of these three threads, the scan frame integration is adjusted based on the distance to the target MAV and its speed.}
    \label{fig:system_overview}
\end{figure*}

\begin{figure}
    \centering
    \includegraphics[width=0.48\textwidth]{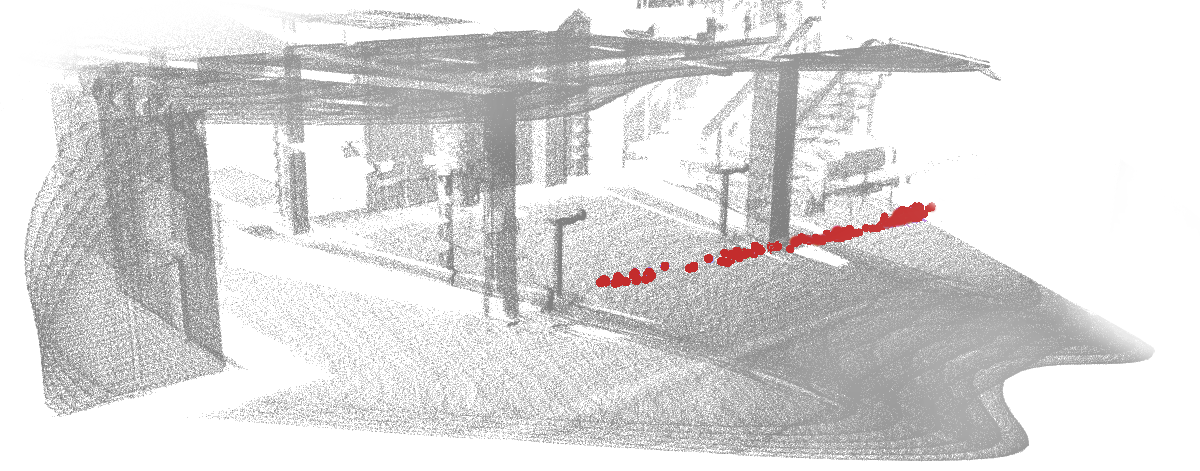}
    \caption{Integration trajectory recovery example}
    \label{fig:trajectory}
\end{figure}

\subsection{Lidar and radar-based MAV detection and tracking}

When the focus is more on detection rather than on accurate localization or tracking of the detected MAV, radar has been proven a robust solution~\cite{fang2018experimental}. Lidars, in any case, have been identified as having big potential for MAV detection and tracking~\cite{hammer2018potential}. 

In summary, while object detection and tracking in point clouds is a relatively mature field, we have found a gap in the literature in terms of optimizing the way these point clouds are generated. In particular, we see most of the current work being focused on processing point clouds, while our objective is to study how we can enhance the performance of a given tracking algorithm by improving the quality of the point cloud data it is fed with. Our focus here is on actively adapting lidar-based perception for detecting and tracking a flying MAV, where the density and size of the point cloud is optimized based on, e.g., the MAVs distance to the lidar sensor, or its speed.

\section{Problem Definition}
\label{sec:definitions}

We consider the problem of tracking a MAV from a ground robot. The ultimate objective is, e.g., to improve the collaboration between the robots and the ability of the MAV to navigate in complex environments aided by the UGV. The rest of this paper delves into the definition, design, and implementation of methods for tracking a single MAV. Nonetheless, these can be extended to multiple MAVs. The main limitation when tracking multiple units is the FoV of the lidar sensors onboard the ground vehicle, and therefore assumptions have to be made to the spatial distribution of the MAVs (always within the FoV of the ground robot). Alternatively, more lidar scanners can be installed to increase the FoV.

\subsection{Rationale}

The majority of 3D laser scanners available to date are multi-channel, rotating lidars. While devices with 64 or 128 vertical channels can provide high angular resolution in both horizontal and vertical dimensions, these high-end devices are not the most common. Moreover, the scanning pattern is in general repetitive, which has benefited from a geometric perspective in terms of data processing but does not enable a higher FoV coverage with longer exposure if the position of the sensor is fixed. New solid-state lidars featuring non-repetitive scan patterns, albeit having more limited FoV, can provide more dense point clouds and often feature longer detection ranges. In particular, we are interested in the possibilities of dynamically adjusting the FoV coverage and density in the point cloud to be processed for detection and tracking. Among the benefits of these new lidars and the possibilities of adaptive scanning rates is also higher resilience against one of the challenges in lidar-based perception: motion-induced distortion~\cite{neuhaus2018mc2slam}. In general, the literature targeting tracking of MAVs using lidar scanners is scarce, and existing methods in point cloud object detection and tracking considering mainly static frames. We aim to define more optimal settings for generating point clouds based on the state (speed and distance to the sensor) of the MAV being tracked.

\subsection{System Overview}

We propose three simultaneous tracking modalities with three processes analyzing point cloud frames resulting in integration times ranging several orders of magnitude. A general view of the multi-modal tracking processes is shown in Fig.~\ref{fig:system_overview}. In more detail, the three modalities are described below:

\begin{enumerate}[(i)]
    \item Adaptive high-frequency tracking. In this first process, sparse point clouds are integrated at frequencies up to 100\,Hz. The MAV is only trackable through a reduced number of points, but we are able to estimate its position and speed with high accuracy. In this process, the MAV is not necessarily recognizable in all processed frames.
    \item Adaptive medium-frequency tracking. The second process operates at frequencies within the range of typical lidar scanners (i.e., 5 to 20\,Hz). The frequency within that same range is dynamically adjusted to optimize the density of the point cloud. At these frequencies, the extracted point cloud representing the MAV is distorted by motion, and thus the localization and speed estimation accuracy is lower. However, this process enables more robust and persistent tracking as the MAV can be recognized in most if not all frames.
    \item Low-frequency trajectory and object validation. The third and last process that runs in parallel to the previous two performs long-term tracking and validates the reconstructed trajectory of the MAV based on predefined dimensional constraints. An illustration of such trajectory reconstruction is shown in Fig.~\ref{fig:trajectory}
\end{enumerate}

\subsection{Formulation}

Let $\mathcal{P}_k(I_{r}^k)$ be the point cloud generated by the lidar with an integration time $I_{r}^k$, and let $\textbf{s}^k_{UGV}$=\{ $\textbf{q}^{k}_{UGV}$, $\dot{\textbf{q}}^{k}_{UGV}$\} be the position and speed defining the state of the UGV at time $k$. We also denote by $\textbf{s}^k_{MAV}$=\{$\textbf{p}^{k}_{MAV}$,$\dot{\textbf{p}}^{k}_{MAV}$\} the position and speed of the MAV. We use discrete steps represented by $k$ owing to the discrete nature of the set of consecutive point clouds. The output of the main tracking algorithm is to extract from $\mathcal{P}_k(I_{r}^k)$ the set of points representing the MAV, which we denote by $\mathcal{P}^{k}_{MAV}$, and to adjust the integration time for the next point cloud, $I^{k}_{HF}, I^{k}_{MF}$.

\begin{algorithm}[t]
    \footnotesize
	\caption{\footnotesize MAV tracking with adaptive scan integration}
	\label{alg:tracking}
	\KwIn{\\ 
	    \begin{tabular}{ll}
	        High- and medium-freq int. rates:  & $\{I^{k-1}_{HF}, I^{k-1}_{MF}\}$ \\[+0.4em]
	        3D lidar point clouds:             & $\{\mathcal{P}_{k}(I_{HF}^{k-1}), \mathcal{P}_{k}(I_{MF}^{k-1})\:\}$ \\[+0.4em]
	        Last known MAV state:              & $ (\textbf{p}^{k-1}_{MAV}, {\dot{\textbf{p}}_{MAV}}^{k-1})$ \\[+0.4em]
	    \end{tabular}
	}
	\KwOut{\\
	    \begin{tabular}{ll}
	        MAV state:      & \{${\textbf{p}^{k}_{MAV}}, \:\dot{\textbf{p}}^{k}_{MAV}\}$ \\[+0.2em]
	        UGV control:    & $\dot{\textbf{q}}^{k}_{UGV}$ \\ 
	        Int. rates:     & $\{I^{k}_{HF}, I^{k}_{MF}\}$
	    \end{tabular}
	}  
	%
	\SetKwFunction{FSub}{$object\_extraction\left(\mathcal{P}, \:I, \:{\textbf{p}^{k-1}_{MAV}}, \:\dot{\textbf{p}}^{k-1}_{MAV}\right)$}
    \SetKwProg{Fn}{Function}{:}{}
    \BlankLine
    \Fn{\FSub}
    {
            \vspace{.3em}
            \begin{tabular}{ll}
                Ground removal:         & $\mathcal{P}^{'} \leftarrow  \mathcal{P};$ \\
                Generate KD Tree:       & $kdtree \leftarrow \mathcal{P}^{'};$ \\
                MAV pos estimation:     & $\hat{\textbf{p}}_{\tiny MAV}^{k} \leftarrow \textbf{p}^{k-1}_{MAV} + \frac{\dot{\textbf{p}}_{MAV}^{k-1}}{I};$ \\[+0.4em]
                MAV points:             & $\mathcal{P}^{k}_{MAV} = KNN(kdtree, \: \hat{\textbf{p}}_{MAV}^{k});$ \\[+0.4em]
                MAV state estimation:   & ${\textbf{p}^{k}_{MAV}} = \frac{1}{\lvert\mathcal{P}^{k}_{MAV}\rvert}\sum_{p\in\mathcal{P}^{k}_{MAV}} p;$ \\
            \end{tabular}
            \KwRet ${\textbf{p}^{k}_{MAV}};$
    }
	\BlankLine
    %
    \tcp{Coarse but persistent tracking}
    \While{new $\mathcal{P}_{k}(I^{k}_{MF})$}{
        ${\textbf{p}^{k'}_{MAV}} =  object\_extraction(\mathcal{P}_{k}(I^{k}_{MF}), I^{k}_{MF}, \textbf{p}^{k-1}_{MAV}, \dot{\textbf{p}}^{k-1}_{MAV});$ \\
    }
    \tcp{Fine-grained estimation}
    \While{new $\mathcal{P}_{k}(I^{k}_{MF})$}{
        ${\textbf{p}^{k''}_{MAV}} =  object\_extraction(\mathcal{P}_{k}(I^{k}_{HF}), I^{k}_{HF}, \textbf{p}^{k-1}_{MAV}, \dot{\textbf{p}}^{k-1}_{MAV});$ \\
        ${\textbf{p}^{k}_{MAV}}, \: {\dot{\textbf{p}}^{k}_{MAV}} \leftarrow estimate\left(\textbf{p}^{k'}_{MAV}, \:\textbf{p}^{k''}_{MAV}\right)$\;
    }
    $\{I^{k}_{HF}, I^{k}_{MF}\} \leftarrow adjust\_integration\_freqs\left(\textbf{p}^{k}_{MAV}, \dot{\textbf{p}}^{k}_{MAV}\right);$ \\[+0.4em]
    $\dot{\textbf{q}}^{k}_{UGV} \leftarrow keep\_within\_FoV\left(\textbf{p}^{k}_{MAV}, \dot{\textbf{p}}^{k}_{MAV}\right);$

    %
    %
    %
\end{algorithm}

\subsection{Adaptive scan integration}

Since we assume that the state of the MAV $(\textbf{p}^{k-1}_{MAV}, {\dot{\textbf{p}}_{MAV}}^{k-1})$ is initially known, the point cloud processing proceeds as follows. First, we perform ground removal based on the known position of the UGV and the last-known altitude of the MAV. We then proceed with finding the nearest neighbor points to a predicted MAV position. This step is repeated for both the high and medium frequency scans, the former one providing a more accurate position estimation while the latter is more persistent in time. Finally, these two estimations are combined, and the results are utilized to adjust the integration rates based on the point cloud density expected for the given distance and speed. The UGV is also controlled to maintain the MAV within the FoV of its lidar. This process is outlined in Algorithm~\ref{alg:tracking}.

\begin{algorithm}[t]
    \footnotesize
	\caption{Trajectory validation}
	\label{alg:validation}
	\KwIn{\\ 
	    \begin{tabular}{ll}
	        Low-freq int. rate:     & $I^{k-1}_{LF}$ \\[+0.4em]
	        3D lidar point cloud:   & $\mathcal{P}_{k}\left(I_{LF}^{k-1}\right)$ \\[+0.6em]
	        MAV state history:      & $\left(\textbf{p}_{MAV}, \:{\dot{\textbf{p}}_{MAV}}\right)$ \\[+0.8em]
	    \end{tabular}
	}
	\KwOut{ 
	    Trajectory validation ($bool$)
	}  
	\BlankLine 
    \While{new $\mathcal{P}_{k}\left(I_{LF}^{k-1}\right)$ }
    {
        \tcp{Generate cubic splines}
        \tcp{with position and speed constraints}
        $\{B_i\} \leftarrow \{\textbf{p}_{MAV}, \:{\dot{\textbf{p}}_{MAV}}\};$\\
        \tcp{Estimate expected point cloud from}
        \tcp{known density at given distance and speed}
        $\hat{\mathcal{P}}_k \leftarrow \left\{\{B_i\}, \:\textbf{p}_{MAV}, \:{\dot{\textbf{p}}_{MAV}}\right\};$ \\
        \tcp{Calculate IoU}
        $IoU = calc\_IoU\left(\mathcal{P}_{k}\left(I_{LF}^{k-1}\right), \hat{\mathcal{P}}_k\:\right);$ \\
        \eIf{ $IoU > th$ }
            {return True}
            {return False}
    }
 
\end{algorithm}

\subsubsection{Trajectory validation}

The main purpose of the low-frequency scan stream is to validate the extracted MAV's trajectory. While the tracking with adaptive scan integration only takes into account the MAV size roughly in terms of distance within which nearest neighbors are looked for, the extracted point cloud is not validated against its known dimensions. This is done when enough points are accumulated into a reconstructed trajectory. As exposed in Algorithm~\ref{alg:validation}, we first perform a cubic spline interpolation based on the history of estimated positions and speeds. To calculate the parameters of the cubic spline, we utilize constraints on the first derivative based on the speed, rather than forcing the first and second derivative to be continuous. Indeed, the acceleration of the MAV can suddenly change. Based on predetermined values of point cloud density as a function of the MAV's distance to the lidar and its speed, we then produce an expected point cloud. We validate the original point cloud given a threshold for the IoU measure with the generated estimate.

\section{Methodology}

\subsection{Experimental platforms}

The experimental platforms consist on a single ground robot and a commercially available Ryze Tello MAV. The ground robot is an EAI Dashgo platform equipped with a Livox Horizon lidar ($81.7\degree\times25.1\degree$ FoV). The lidar is able to output scanned pointcloud up to 100\,Hz, featuring a non-repetitive pattern. A pair of ultra-wideband (UWB) transceivers is used to obtain a single range between the robot and the MAV at frequencies ranging from 10\,Hz to 100\,Hz. The UWB ranging is only used in aiding the manual validation of the extracted trajectory in places where there was no external positioning system. In the future, it could be incorporated as part of the tracking algorithm as well, as is becoming increasing adopted in multi-robot systems~\cite{shule2020uwb, almansa2020autocalibration}. 

\begin{figure}
    \centering
    \includegraphics[width=.48\textwidth]{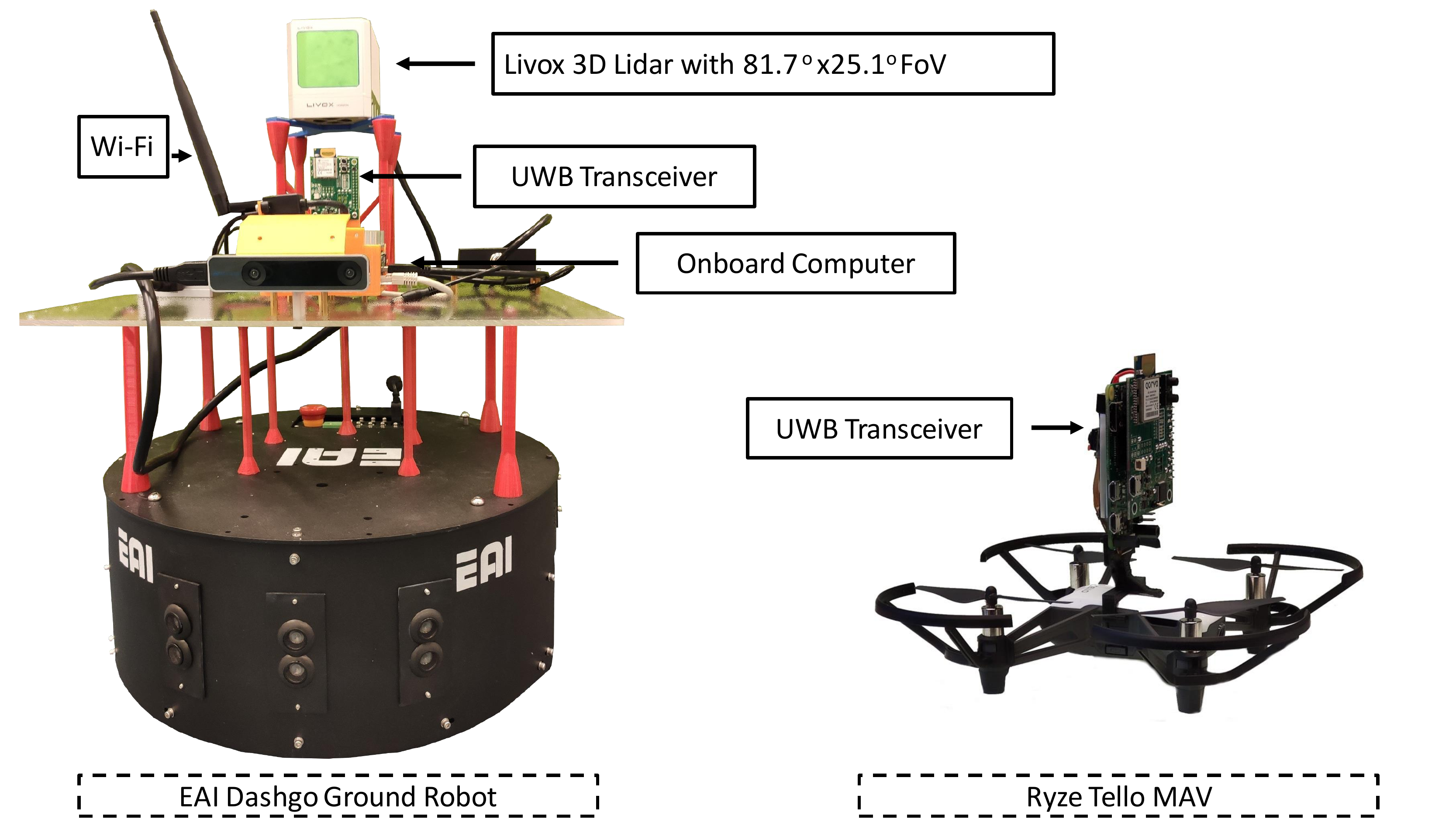}
    \caption{Ground robot and MAV utilized in the experiments.}
    \label{fig:robots}
\end{figure}

\subsection{Software}

The system has been implemented using ROS Melodic under Ubuntu 18.04. The algorithms are running in the main computer onboard the ground robot. The computer runs the robot's driver\footnote{\url{https://github.com/TIERS/dashgo-d1-ros}}, the Tello MAV driver\footnote{\url{https://github.com/TIERS/tello-driver-ros}}, the Livox lidar driver\footnote{\url{https://github.com/Livox-SDK/livox_ros_driver}}, and our open-source MAV tracking package\footnote{\url{https://github.com/TIERS/adaptive-lidar-tracking}}. The latter is a multi-threaded node able to process the different point clouds in real time. The point cloud library (PCL)~\cite{rusu20113d} is utilized to extract the position of the MAV from the lidar's point cloud.

\subsection{Metrics}

Owing to the lack of an accurate external positioning system such as a motion capture system, our focus is instead on measuring the performance of the tracking at different scan integration rates and manually validating the overall trajectory. The experimental flights are carried out in large indoor halls with multiple columns and objects, as shown in Fig.~\ref{fig:trajectory}. Another set of experiments is carried out in a small flying area where an external UWB positioning system was available and used to fly the MAV over a predefined trajectory. A characterization on the accuracy of such system can be found in~\cite{queralta2020uwbbased}.

\begin{figure*}
    \centering
    \setlength\figureheight{0.33\textwidth}
    \setlength\figurewidth{\textwidth}
    \footnotesize{\input{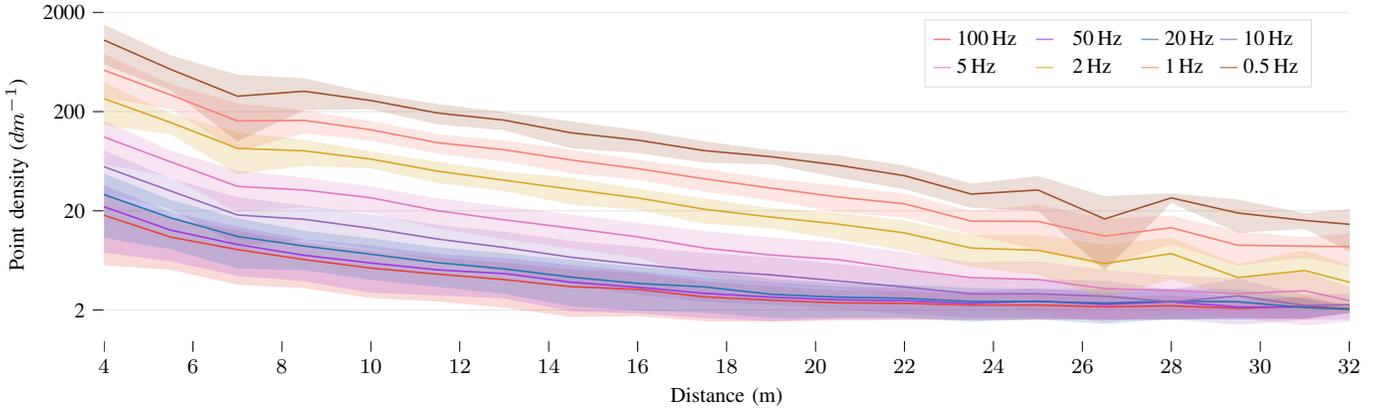}} \\
    \caption{Density of the point cloud representing the MAV based on the distance to the lidar scanner and the scanning frequency.}
    \label{fig:distances}
\end{figure*}
 
 \begin{figure*}
    \centering
    \setlength\figureheight{0.33\textwidth}
    \setlength\figurewidth{\textwidth}
    \footnotesize{\input{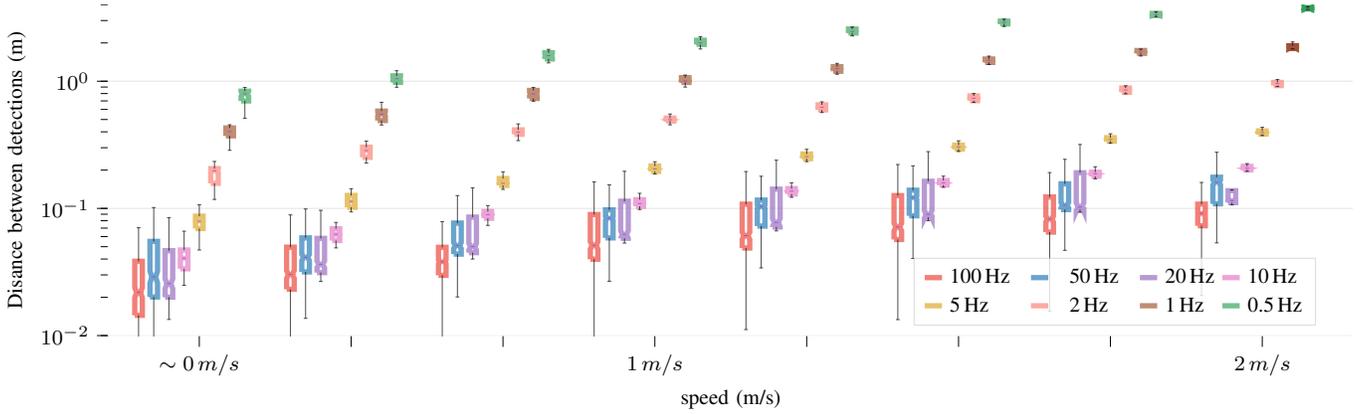}}
    \caption{Distance between consecutive MAV detections based on its speed and the lidar's scanning frequency.}
    \label{fig:speeds}
\end{figure*}

\begin{figure}
    \centering
    \includegraphics[width=0.42\textwidth]{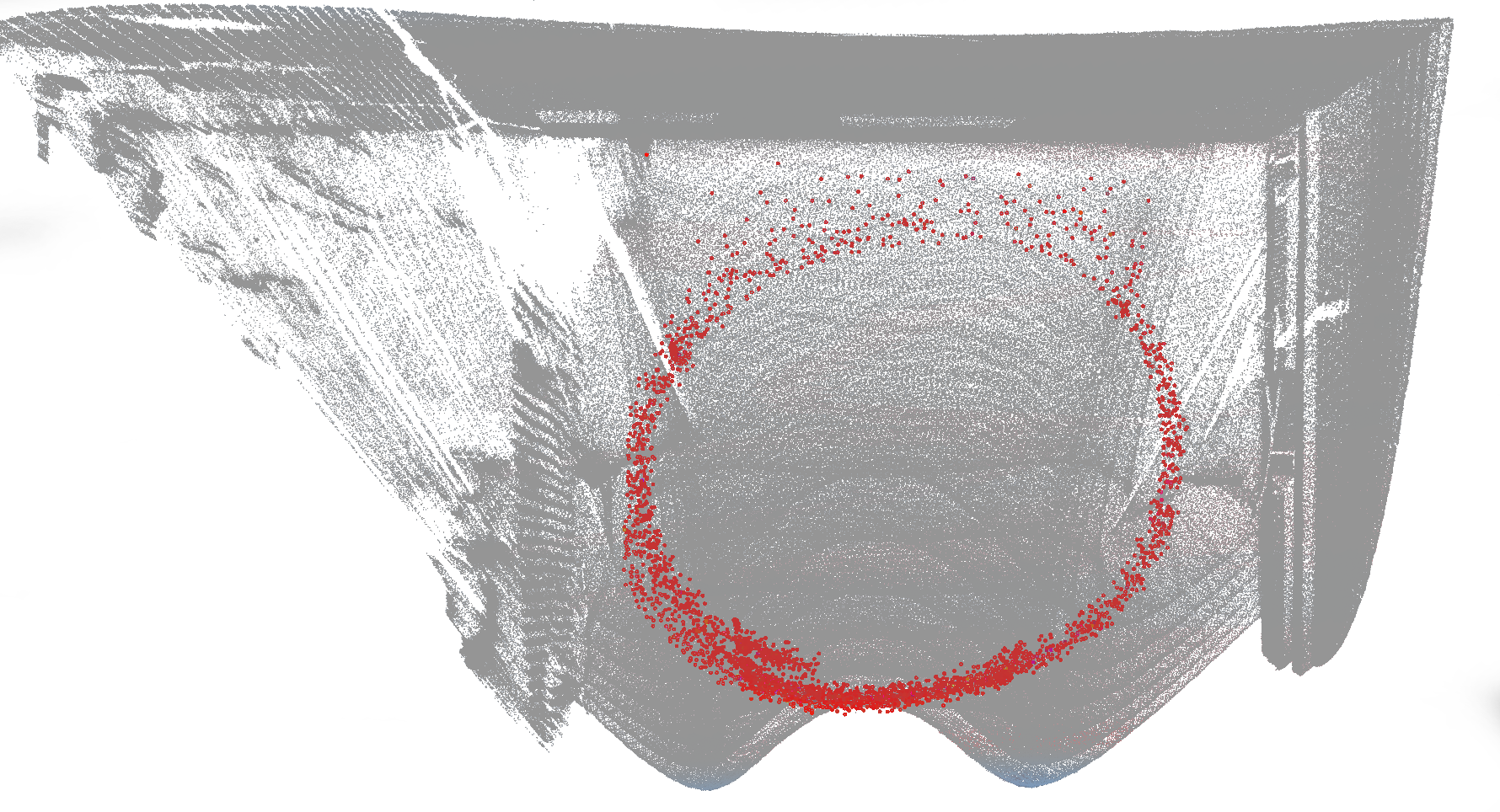}
    \caption{Accumulated point cloud for the circular trajectory.}
    \label{fig:basement_pointcloud}
\end{figure}

\begin{figure}
    \centering
    \setlength\figureheight{0.2\textwidth}
    \setlength\figurewidth{0.5\textwidth}
    \scriptsize{\input{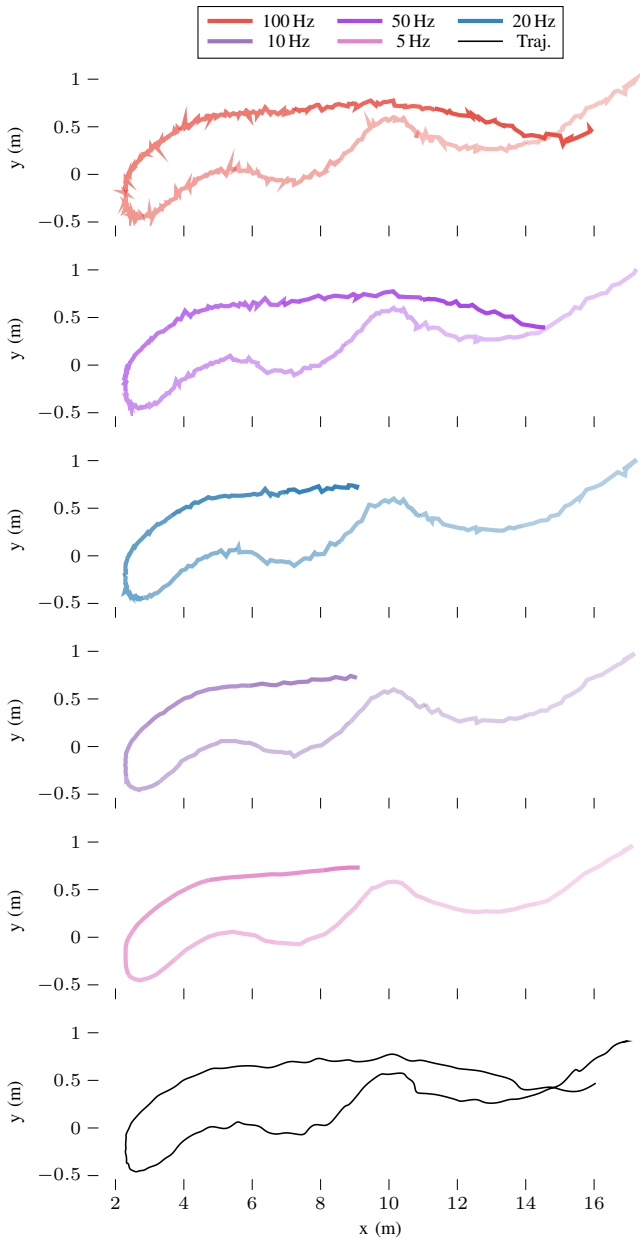}}
    \caption{Estimated trajectories at different frequencies and with adaptive approach (top five plots), and trajectory estimated from our algorithm (bottom plot).}
    \label{fig:natura}
\end{figure}

\begin{figure*}
    \centering
    \setlength\figureheight{0.25\textwidth}
    \setlength\figurewidth{0.3\textwidth}
    \scriptsize{\input{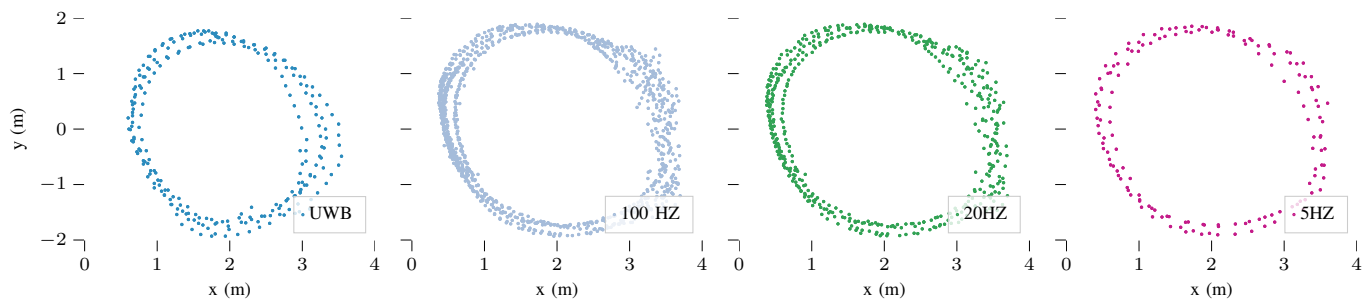}}
    \caption{Reference trajectory (UWB) and estimated positions at different fixed frequencies.}
    \label{fig:basement}
\end{figure*}

\section{Experimental Results}\label{sec:methodology}

In this section we report on the experimental results. The experimental results consist mainly on flights in two different indoor environments and different conditions. 

\subsection{Adaptive scan integration}

The first objective of our experiments was to assess the tracking performance at different scan frequencies in order to better model the adaptiveness of our algorithm. In order to adapt the scanning frequency to optimize the tracking performance, key parameters are the point cloud density at different distances and the reliability of the detections at different speeds.

The point cloud density for different scanning frequencies as a function of the distance between the lidar and the MAV is shown in Fig.~\ref{fig:distances}. This measure refers only to the density of the points representing the MAV and not the overall density including the rest of the scene. The darker lines represent the average point cloud density, while the band with higher transparency represents the values within the standard deviation. The size of the Tello MAV is about 500 cubic centimeters. Based on our experiments, reliable tracking at high speeds can be achieved with at least 4 points, while we require at least 20 points at medium scanning frequency. This, however, only applies in free space. As can be seen in Fig.~\ref{fig:basement_pointcloud}, significant noise appears in the point cloud between the MAV and walls in the environment when flying nearby. We discuss further this issue at the end of this section.

In terms of the tracking performance based on the speed, we plot in Fig.~\ref{fig:speeds} the distance between consecutive detections at different scanning frequencies. The results in this particular figure cannot be directly utilized to model the adaptive nature of our tracking algorithm. Nonetheless, they can be leveraged to better understand what are the speed limits under which given scanning frequencies do not provide the expected distance between detection that can be inferred from the MAV speed and the scan frequency.

The results included in Fig.~\ref{fig:distances} and Fig.~\ref{fig:speeds} have been obtained flying the MAV in a long, straight corridor with a length of about 35\,m. The MAV was flying mostly in straight lines and the speed was estimated using both visual odometry and the position history extracted from the lidar data in a partially manual manner.

\subsection{Qualitative trajectory validation}

In order to validate the performance of the tracking algorithm and better understand the limitations of our tracking approach at different scanning frequencies, we compare two different types of trajectories. Owing to the lack of a system to obtain ground truth (e.g., a motion capture system), we provide qualitative analysis for one of the trajectories and compare it with a UWB positioning system in the other one.

First, we test the tracking algorithm through a trajectory where the MAV flies in a large open area at distances from 2\,m to over 17\,m far from the lidar scanner and variable speeds. In this scenario, the analysis is mostly qualitative, with the trajectories shown in Fig.~\ref{fig:natura}. However, the UWB ranging data and the lidar data has been both manually confirmed, so the maximum positioning error along the track is at worst around 20\,cm. Qualitatively, the main results from this experiment are the ability of the tracking algorithm to keep track of the MAV over changes in speed, direction, and at longer distances. The figure only shows frequencies equal to or above 5\,Hz because at lower scanning frequencies the speed estimation was highly inaccurate during the early stages of the flight. We can see that only at the highest frequency we are able to track the MAV along the completed trajectory, while the trajectory itself is noisier. The higher level of error when estimating the MAV position is due to a lower number of points being detected, which can correspond to different parts of the MAV in consecutive scans. The last subplot shows the overall estimated trajectory where our algorithm has combined the different scanning frequencies to obtain the smoothness of the medium frequencies and the performance of the higher frequencies. The trajectory also employs the cubic spline interpolation from the validation algorithm.

Second, we perform a continuous flight with a predefined circular trajectory in a small flying arena where the UWB positioning system is available. The results for this flight are shown in Fig.~\ref{fig:basement}. The leftmost plot shows the reference position. However, it is worth noticing that the accuracy of the lidar, of around 2\,cm for distances smaller than 20\,m, is higher than the average accuracy of 10 to 15\,cm in the UWB positioning system. Therefore, the trajectory is mere as a reference and only a qualitative discussion is possible with these results. In any case, owing to the continuous change in the speed of the MAV, which is a prior unknown to the tracking algorithm, again only at frequencies equal or over 5\,Hz are we able to track the MAV. Nonetheless, at 5\,Hz the tracking stops before the fourth revolution is completed, and persistent tracking is only possible when higher frequencies are taken into account.

\subsection{Discussion}

We have shown in this section qualitative results that show the performance of the adaptive tracking algorithm and the same approach applied only to specific scanning frequencies. From both sets of experiments, the main conclusion is that the adaptive approach is able to accommodate a wider variety of scenarios. We have been able to put together the flexibility of high-speed tracking with the robustness of medium frequencies, avoiding the frequent errors of the former, and the lower tracking capacity of the latter is more challenging conditions.

One key limitation when tracking MAVs, as visualized in the circular trajectory experiments, is the low density of the point cloud and the inability to tell the difference between the MAV's points and lidar noise. This is also due to the low reflectively of the MAV, and there is thus the potential for mitigation with more reflective surfaces that could aid in separating the sparse MAV point cloud from the lidar noise originated due to near objects. As we can see in Fig.~\ref{fig:basement_pointcloud}, the point cloud density near the rear wall is very sparse in some areas, therefore being unable to reconstruct a robust trajectory as there are multiple options available that would meet the dynamics and dimensional constraints of the MAV.

\section{Conclusion}
\label{sec:conclusions}

We have presented a set of methods for detecting and tracking MAVs that are deployed from ground robots, assuming that the initial position is known. The focus has been on the introduction of a novel adaptive lidar scan integration method that enables more accurate MAV localization with high-frequency scans, robust and persistent tracking with longer frame integration times, and trajectory validation with low-frequency analysis. Experimental results from different settings confirm the better suitability of the different integration times for different scenarios or MAV behaviour, with our adaptive tracking being able to consistently track a MAV in places where a constant lidar scan frequency cannot. Finally, with an additional method to validate the trajectory based on the known shape and size of the MAV, we are able to confirm that the object being tracked meets the dimensional constraints.

In future works, we will explore the integration of lidar-based tracking into the navigation of the MAV, and the integration of onboard state estimation at the MAV into the tracking algorithm.

\section*{Acknowledgment}

This research work is supported by the Academy of Finland's AutoSOS project (Grant No. 328755) and Finnish Foundation for Technology Promotion.

\bibliographystyle{IEEEtran}
\bibliography{bibliography}

\end{document}